\documentclass{article}

\usepackage[preprint]{neurips_2026}

\usepackage[utf8]{inputenc}
\usepackage[T1]{fontenc}
\usepackage{hyperref}
\usepackage{url}
\usepackage{booktabs}
\usepackage{amsfonts}
\usepackage{nicefrac}
\usepackage{microtype}
\usepackage{xcolor}
\usepackage{algorithm}

\usepackage{graphicx}
\usepackage{amsmath}

\usepackage{algorithmicx}
\usepackage{algpseudocode}

\title{
Comparative Evaluation of Machine Learning Approaches for Minority-Class Financial Distress Prediction Under Class Imbalance Constraints
}

\author{
Karan Sehgal \\
Kent Business School \\
University of Kent \\
Canterbury, United Kingdom \\
\texttt{K.Sehgal@kent.ac.uk}
\and
Khawar Naveed Bhatti \\
Kent Business School \\
University of Kent \\
Canterbury, United Kingdom \\
\texttt{K.Bhatti@kent.ac.uk}
}

\begin{document}

\maketitle

\begin{abstract}

Financial distress prediction remains a significant challenge in enterprise risk analysis due to the highly imbalanced nature of real-world financial datasets, where bankrupt or distressed firms typically constitute only a small minority of observations. Under such conditions, conventional classification models may achieve strong aggregate accuracy while demonstrating insufficient sensitivity toward minority-class distress events.

This paper presents a comparative evaluation of classical statistical methods, ensemble learning approaches, and exploratory neural models for minority-class financial distress prediction under severe class imbalance constraints. The study adopts a CRISP-DM-oriented machine learning workflow incorporating structured preprocessing, feature filtering, imbalance mitigation using the Synthetic Minority Oversampling Technique (SMOTE), comparative model optimisation, and explainability-oriented evaluation.

Comparative experimentation was conducted across multiple model families including Logistic Regression, Random Forest, AdaBoost, XGBoost, CatBoost, and LightGBM, alongside exploratory temporal modelling approaches. Model performance was evaluated using minority-class precision, recall, F1-score, and ROC-AUC metrics, with particular emphasis placed on recall due to the asymmetric operational risk associated with false negatives in enterprise financial environments.

To support interpretability and governance-oriented analysis, SHAP-based explainability methods were additionally applied to investigate feature-level contributions toward bankruptcy-event prediction. The broader objective of the study is to support reproducible, interpretable, and imbalance-aware machine learning workflows for enterprise financial distress prediction under high-risk operational conditions.

\end{abstract}

\section{Introduction}

Financial distress prediction is an important classification problem in financial risk management, insolvency monitoring, enterprise governance systems, and operational decision-support environments. Early identification of financially distressed firms may assist organisations in reducing exposure to insolvency-related losses, improving strategic risk oversight, and supporting more reliable enterprise financial planning. However, practical implementation of predictive systems for financial distress remains challenging due to the rarity of bankruptcy events within real-world financial datasets.

In many enterprise and public financial datasets, bankrupt firms represent only a small proportion of total observations, producing a severely imbalanced classification environment. Under such conditions, conventional machine learning models may become biased toward majority non-bankrupt observations, resulting in inflated aggregate accuracy while demonstrating poor sensitivity toward minority-class distress events. This issue is particularly significant within financial risk environments because false negatives may delay intervention and increase downstream operational and financial exposure.

Traditional financial distress prediction methods such as the Altman Z-score \cite{altman1968}, Ohlson O-score \cite{ohlson1980}, and Zmijewski model \cite{zmijewski1984} have historically provided interpretable statistical frameworks for insolvency estimation. These approaches remain influential due to their methodological simplicity and interpretability, although they may demonstrate limitations under enterprise environments characterised by nonlinear financial interactions, heterogeneous operational conditions, and evolving market behaviour.

More recent machine learning approaches including XGBoost \cite{chen2016xgboost}, LightGBM \cite{ke2017lightgbm}, and CatBoost \cite{prokhorenkova2018catboost} offer improved flexibility for modelling nonlinear financial relationships and complex feature dependencies within structured enterprise datasets. Ensemble learning architectures have increasingly demonstrated strong predictive capability across bankruptcy prediction and financial risk modelling tasks involving heterogeneous financial indicators.

Despite advances in predictive modelling, several important challenges remain relevant within enterprise-oriented financial machine learning systems. Severe class imbalance requires explicit imbalance-aware optimisation strategies to improve minority-event sensitivity during model training and evaluation. In parallel, increasing enterprise governance and regulatory expectations require predictive systems to support explainability, transparency, reproducibility, and operational auditability rather than functioning as opaque black-box classifiers.

This study evaluates multiple machine learning approaches for minority-class financial distress prediction under severe class imbalance conditions using SMOTE \cite{chawla2002smote} and SHAP explainability analysis \cite{lundberg2017unified}. The study adopts a CRISP-DM-oriented experimentation framework incorporating preprocessing, imbalance mitigation, comparative model evaluation, and governance-oriented explainability analysis across multiple statistical, ensemble, and exploratory temporal modelling approaches.

The broader objective of the work is not solely predictive optimisation, but the development of a reproducible and interpretable machine learning evaluation framework capable of supporting trustworthy enterprise financial risk analysis under highly imbalanced operational conditions.

\section{Literature Review}

\subsection{Classical Financial Distress Prediction Models}

Financial distress prediction has historically relied on statistical scoring models derived from accounting and financial ratio analysis. One of the most widely recognised approaches is the Altman Z-score model \cite{altman1968}, which applies multiple discriminant analysis to estimate bankruptcy risk using financial ratios associated with liquidity, profitability, leverage, and solvency.

Subsequent models including the Ohlson O-score \cite{ohlson1980} and Zmijewski model \cite{zmijewski1984} introduced logistic and probit-based statistical approaches for insolvency estimation. These methods remain important due to their interpretability, methodological simplicity, and relatively low computational complexity.

However, classical statistical models may demonstrate limitations under modern enterprise environments characterised by nonlinear financial interactions, heterogeneous reporting structures, noisy operational indicators, and evolving market conditions. Many traditional approaches additionally rely on assumptions involving linear separability and relatively stable financial behaviour, which may reduce predictive flexibility in complex real-world classification environments.

Despite these limitations, classical financial distress prediction models continue to provide important baseline frameworks for comparative evaluation and remain influential within financial risk modelling literature due to their interpretability and historical significance.

Subsequent models including the Ohlson O-score \cite{ohlson1980} and Zmijewski model \cite{zmijewski1984} introduced logistic and probit-based statistical approaches for insolvency estimation. These methods remain important due to their interpretability and relatively low computational complexity.

However, classical statistical models may demonstrate limitations under modern enterprise environments characterised by nonlinear financial interactions, heterogeneous reporting structures, and noisy operational indicators.

\subsection{Machine Learning Approaches for Financial Distress Prediction}

Machine learning methods have increasingly been applied to financial distress prediction due to their ability to model nonlinear relationships, heterogeneous financial behaviour, and complex feature dependencies within structured enterprise datasets. Unlike traditional statistical scoring approaches, machine learning architectures are capable of capturing higher-order interactions between financial indicators under dynamic operational conditions.

Ensemble learning approaches including Random Forest, AdaBoost, XGBoost \cite{chen2016xgboost}, CatBoost \cite{prokhorenkova2018catboost}, and LightGBM \cite{ke2017lightgbm} have demonstrated strong predictive capability across tabular financial classification tasks involving bankruptcy prediction, fraud detection, and enterprise risk modelling.

Random Forest models improve robustness through bootstrap aggregation and feature randomisation, thereby reducing variance and improving generalisation performance under noisy financial environments. Gradient-boosting architectures including XGBoost, CatBoost, and LightGBM further improve predictive performance through iterative optimisation of residual classification errors using sequential ensemble construction strategies.

These boosting-based approaches are particularly well suited to structured enterprise datasets due to their ability to model nonlinear feature dependencies, manage heterogeneous variable distributions, and maintain comparatively strong predictive performance under imbalance-aware optimisation conditions.

Neural approaches including Long Short-Term Memory (LSTM) networks have additionally been explored for modelling temporal financial behaviour within sequential financial datasets. Such architectures may improve representation of evolving financial conditions by capturing temporal dependencies across historical enterprise indicators. In parallel, statistical forecasting approaches including ARIMA and SARIMA remain relevant baseline techniques for structured temporal financial analysis.

Despite their predictive advantages, machine learning approaches for financial distress prediction remain highly sensitive to preprocessing quality, imbalance mitigation procedures, evaluation design, and explainability considerations. Consequently, reproducibility, minority-class sensitivity, and governance-oriented evaluation frameworks remain important considerations within enterprise financial machine learning systems.

\subsection{Class Imbalance in Financial Prediction}

Class imbalance represents a major challenge in financial distress prediction systems because bankrupt observations are typically rare relative to non-distressed enterprise cases. Under highly imbalanced classification conditions, machine learning models may become disproportionately biased toward majority-class observations, resulting in artificially inflated aggregate accuracy while demonstrating poor sensitivity toward minority-event detection.

This issue is particularly significant within enterprise financial risk environments because false negatives may delay intervention, reduce insolvency visibility, and increase downstream operational and financial exposure. Consequently, imbalance-aware optimisation and evaluation procedures are considered critical components of reliable financial distress prediction workflows.

Oversampling approaches such as the Synthetic Minority Oversampling Technique (SMOTE) \cite{chawla2002smote} are commonly used to improve minority-class representation during model optimisation. SMOTE generates synthetic minority-class observations through nearest-neighbour interpolation between existing minority samples, thereby reducing majority-class dominance during training procedures.

Prior research has demonstrated that imbalance-aware preprocessing strategies may improve minority-class recall, F1-score performance, and bankruptcy-event sensitivity across financial classification environments. However, oversampling procedures may additionally introduce precision-recall trade-offs and potential overfitting behaviour if synthetic sample generation is not carefully controlled.

As a result, imbalance-aware financial machine learning systems require evaluation frameworks that prioritise minority-event sensitivity, reproducibility, and operational reliability rather than relying exclusively on aggregate accuracy metrics. These considerations are particularly important within enterprise-oriented financial prediction systems where minority-event detection carries disproportionate operational significance.

\subsection{Explainability and Governance in Financial Machine Learning}

As machine learning systems become increasingly integrated into enterprise decision-support environments, explainability and governance considerations have become significantly more important. Financial prediction systems may influence operational decisions involving insolvency monitoring, credit risk analysis, enterprise risk management, regulatory reporting, and strategic financial oversight. Consequently, predictive models require not only acceptable performance, but also sufficient interpretability and transparency to support auditability, governance review, and operational trust.

Post-hoc explainability approaches such as SHAP \cite{lundberg2017unified} provide feature-attribution mechanisms for understanding predictive model behaviour and improving transparency within complex machine learning systems. SHAP estimates the contribution of individual features toward predictive outcomes using cooperative game-theoretic principles, enabling both global and local interpretation of model behaviour across varying financial conditions.

Within enterprise-oriented financial environments, explainability mechanisms are particularly important because high-performing black-box models may otherwise introduce governance challenges associated with accountability, validation, traceability, and regulatory oversight. Feature-attribution analysis allows practitioners to identify the financial indicators most strongly influencing bankruptcy-event prediction and supports improved interpretability of model-assisted decision processes.

Beyond predictive interpretation alone, explainability contributes toward broader objectives involving reproducibility, governance-oriented model validation, trustworthy AI evaluation, and operational auditability. In high-stakes financial systems, these considerations are increasingly important as organisations seek to deploy machine learning workflows capable of supporting transparent, interpretable, and reliable enterprise decision-support infrastructure.

\section{Methodology}

This study adopts an applied machine learning evaluation framework for minority-class financial distress prediction under severe class imbalance conditions. The methodology was designed to support reproducible comparative analysis across multiple statistical, ensemble, and exploratory temporal modelling approaches within enterprise financial risk environments.

The overall workflow follows a CRISP-DM-oriented structure incorporating preprocessing, imbalance mitigation, model optimisation, comparative evaluation, and explainability analysis stages. Particular emphasis was placed on minority-class sensitivity, evaluation reliability, interpretability, and governance-oriented machine learning considerations due to the operational significance of false negatives in enterprise financial distress environments.

The engineering objective of the methodology was not solely predictive optimisation, but the development of an interpretable and reproducible experimentation pipeline capable of supporting trustworthy financial risk evaluation under highly imbalanced classification conditions. To support this objective, structured preprocessing workflows, controlled oversampling procedures, cross-validation strategies, and explainability-oriented evaluation mechanisms were incorporated throughout the modelling lifecycle.

The methodology additionally prioritised transparency and operational auditability by integrating feature-attribution analysis and reproducible experimentation procedures across all major stages of the workflow. This broader systems-oriented approach aligns with emerging enterprise requirements surrounding explainable AI, trustworthy machine learning infrastructure, and governance-aware intelligent decision-support systems.

\begin{algorithm}
\caption{Imbalance-Aware Financial Distress Prediction Workflow}

\begin{algorithmic}[1]

\State Input: Financial dataset \(D\)

\State Preprocess dataset
\State Handle missing values
\State Apply feature scaling
\State Perform train-test split
\State Apply SMOTE on training data
\State Train ensemble classifiers
\State Evaluate using Recall, F1-score, and ROC-AUC
\State Apply SHAP explainability analysis
\State Compare model performance

\State Output: Best-performing predictive workflow

\end{algorithmic}
\end{algorithm}

\subsection{Research Design}

This study follows an applied machine learning evaluation framework for minority-class financial distress prediction under severe class imbalance conditions.

The research methodology was structured using a CRISP-DM-oriented workflow incorporating data preprocessing, feature filtering, imbalance mitigation, model training, comparative evaluation, and explainability analysis.

The primary objective of the study was to evaluate the effectiveness of multiple machine learning approaches for improving minority-class sensitivity within highly imbalanced enterprise financial datasets.

Particular emphasis was placed on evaluation reliability, reproducibility, and interpretability due to the operational significance of false negatives in financial distress prediction environments.

The workflow incorporated structured preprocessing pipelines, imbalance-aware optimisation procedures, comparative model benchmarking, and post-hoc explainability analysis using SHAP-based feature attribution methods.

Rather than proposing a novel algorithmic architecture, the study is positioned as an applied engineering evaluation intended to assess the practical behaviour of ensemble learning and exploratory temporal modelling approaches under severe imbalance constraints.

The overall research design additionally emphasised governance-oriented machine learning considerations including auditability, reproducibility, transparency, and interpretable decision-support evaluation within enterprise financial risk environments.

\subsection{Dataset Description}

The dataset used in this study consisted of structured financial and operational indicators associated with enterprise bankruptcy prediction obtained from publicly available financial records hosted through the Taiwan Economic Journal dataset and related Kaggle research repositories. The dataset incorporated firm-level financial variables associated with liquidity, profitability, leverage, retained earnings, solvency, and operational performance.

The dataset contained approximately 6,819 firm observations and 96 financial attributes, with bankruptcy instances representing approximately 1.8\% of total observations. This produced a severely imbalanced classification environment in which minority-class bankruptcy events were substantially underrepresented relative to non-distressed firms.

The classification problem was formulated as a binary prediction task in which firms were categorised as either bankrupt/distressed or non-bankrupt. Due to the rarity of minority-class events, the dataset presented significant challenges associated with imbalance-aware optimisation, evaluation reliability, and minority-event sensitivity during model training.

Prior to preprocessing and imbalance mitigation procedures, the dataset was partitioned using an 80/20 train-test split to preserve evaluation integrity and reduce information leakage between training and testing environments. Missing values were handled using median-imputation strategies where applicable, while numerical features were normalised using StandardScaler preprocessing.

The dataset structure and imbalance characteristics provided an appropriate evaluation environment for investigating reproducible and interpretable machine learning workflows under enterprise financial risk conditions involving severe minority-class sparsity.

\subsection{Data Preprocessing}

Preprocessing procedures were implemented to improve data quality, reduce feature redundancy, and support model stability across comparative experimentation workflows. The preprocessing pipeline was additionally designed to improve reproducibility, minimise optimisation instability, and support evaluation consistency under severe imbalance conditions.

Missing values were handled using median-imputation strategies where applicable in order to reduce sensitivity to incomplete financial observations while preserving overall dataset structure. Numerical financial indicators were normalised using StandardScaler preprocessing to improve comparability across variables exhibiting differing scales, magnitudes, and statistical distributions.

Correlation-based feature filtering was additionally performed to reduce multicollinearity and eliminate highly redundant financial indicators. Features demonstrating excessive pairwise correlation were selectively removed to reduce duplicated informational influence during optimisation and improve downstream model robustness across ensemble learning architectures.

The preprocessing workflow additionally incorporated controlled dataset partitioning and imbalance-aware preparation procedures prior to oversampling stages. These controls were introduced to preserve evaluation integrity and reduce the likelihood of information leakage between training and testing environments.

Overall, the preprocessing pipeline was structured as part of a broader reproducibility-oriented engineering workflow intended to support transparent, stable, and interpretable comparative machine learning evaluation within enterprise financial distress prediction environments.

\subsection{Imbalance Mitigation Using SMOTE}

To address severe class imbalance, the Synthetic Minority Oversampling Technique (SMOTE) \cite{chawla2002smote} was applied to the training data as part of the imbalance-aware optimisation workflow.

SMOTE generates synthetic minority-class observations through nearest-neighbour interpolation between existing minority samples, thereby improving representation of bankrupt observations during model optimisation. The objective of this procedure was to reduce majority-class dominance during training and improve sensitivity toward minority-event detection under highly imbalanced financial classification conditions.

Synthetic minority samples were generated using nearest-neighbour interpolation according to:

\begin{equation}
x_{new} = x_i + \lambda(x_{nn} - x_i)
\end{equation}

where \(x_i\) represents an existing minority-class observation, \(x_{nn}\) denotes one of its nearest minority neighbours, and \(\lambda \in [0,1]\) is a randomly sampled interpolation coefficient.

Oversampling procedures were applied exclusively to the training partition in order to preserve evaluation integrity and avoid information leakage into holdout testing environments. This separation was considered particularly important for maintaining reliable comparative evaluation under reproducibility-oriented experimentation workflows.

\begin{figure}[h]
\centering
\includegraphics[width=0.90\linewidth]{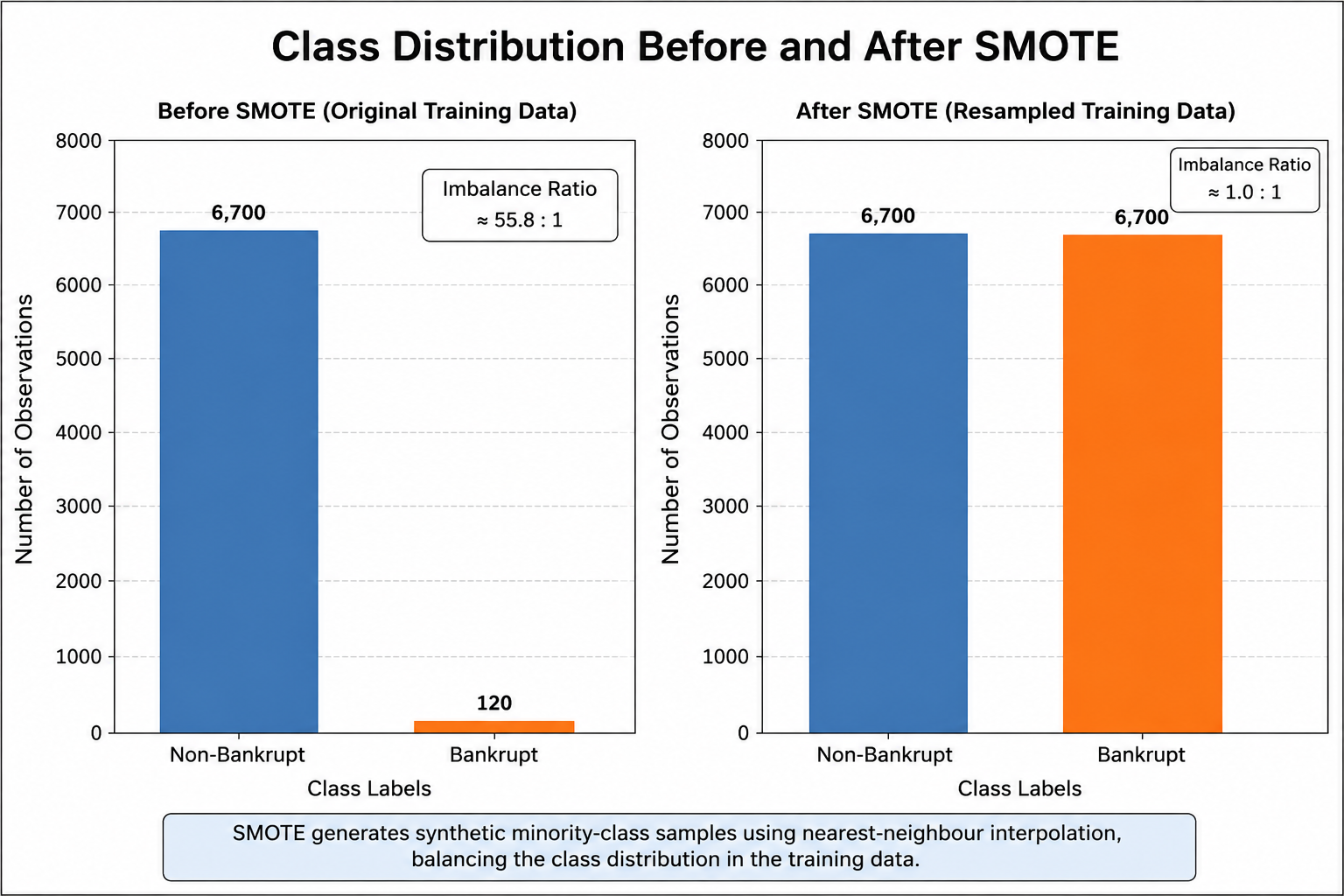}
\caption{Class distribution before and after SMOTE-based oversampling. Minority-class bankruptcy observations were synthetically augmented to reduce majority-class dominance during model optimisation.}
\label{fig:smote_distribution}
\end{figure}

The imbalance-aware pipeline was designed to improve minority-class recall and detection sensitivity without altering the underlying distribution of the evaluation dataset. Given the asymmetric operational risk associated with false negatives in enterprise financial distress environments, imbalance mitigation was treated as a central component of the overall modelling and evaluation framework.

Although SMOTE contributed positively toward minority-class representation during optimisation, the workflow additionally considered potential precision-recall trade-offs associated with synthetic oversampling under sparse minority-event conditions. Consequently, evaluation procedures prioritised balanced analysis across recall, F1-score, ROC-AUC, and broader model reliability considerations rather than relying solely on aggregate accuracy metrics.

\subsection{Machine Learning Models}

A comparative evaluation was conducted across multiple statistical, ensemble, and exploratory neural modelling approaches in order to assess predictive behaviour under severe class imbalance conditions.

The evaluated models included:

\begin{itemize}
    \item Logistic Regression
    \item Random Forest
    \item AdaBoost
    \item XGBoost
    \item CatBoost
    \item LightGBM
\end{itemize}

Logistic Regression was included as a baseline statistical classifier for comparative evaluation against nonlinear ensemble architectures. Despite its interpretability and computational simplicity, linear classification behaviour may demonstrate limitations under highly heterogeneous financial environments characterised by complex feature interactions and minority-class sparsity.

Random Forest models were evaluated due to their robustness under structured tabular environments and their ability to model nonlinear feature relationships through bootstrap aggregation and feature randomisation mechanisms. Ensemble tree approaches additionally provide comparatively stable generalisation performance under noisy financial conditions.

Gradient-boosting architectures including XGBoost, CatBoost, and LightGBM were selected due to their strong predictive capability within enterprise-oriented tabular machine learning workflows. These approaches iteratively optimise residual classification errors through sequential ensemble construction and have demonstrated strong performance across financial classification and risk prediction tasks involving nonlinear feature dependencies.

Hyperparameter optimisation procedures were performed using cross-validation strategies where computationally feasible in order to improve evaluation consistency and reduce overfitting risk across comparative experiments.

In addition to ensemble learning approaches, exploratory temporal modelling techniques including Long Short-Term Memory (LSTM), ARIMA, and SARIMA architectures were investigated to assess sequential financial signal behaviour and temporal forecasting potential under structured historical financial conditions.

However, the primary focus of this study remained imbalance-aware tabular classification using ensemble learning approaches due to their stronger stability, interpretability, and predictive performance under severe minority-class imbalance environments. The comparative modelling framework was therefore designed not only to evaluate predictive effectiveness, but also to investigate broader considerations involving reproducibility, operational reliability, explainability, and governance-oriented machine learning evaluation within enterprise financial risk systems.

\subsection{Evaluation Framework}

The evaluation framework prioritised minority-class detection performance due to the asymmetric cost associated with false negatives in financial distress prediction environments.

Conventional classification accuracy was not treated as the primary evaluation metric because highly imbalanced datasets may produce misleadingly high aggregate accuracy despite poor minority-class sensitivity.

The following evaluation metrics were therefore incorporated:

\begin{itemize}
    \item Precision
    \item Recall
    \item F1-score
    \item ROC-AUC
\end{itemize}

The evaluation metrics were formally defined as follows:

\begin{equation}
Precision = \frac{TP}{TP + FP}
\end{equation}

\begin{equation}
Recall = \frac{TP}{TP + FN}
\end{equation}

\begin{equation}
F1 = 2 \cdot \frac{Precision \cdot Recall}{Precision + Recall}
\end{equation}

where \(TP\), \(FP\), and \(FN\) denote true positives, false positives, and false negatives respectively.

ROC-AUC was additionally used to evaluate discriminatory capability across varying classification thresholds and to assess model robustness under severe imbalance conditions.

Particular emphasis was placed on minority-class recall because failure to detect financially distressed firms may contribute toward increased operational and financial exposure within enterprise risk systems.

Cross-validation and holdout evaluation procedures were additionally incorporated to improve robustness, reduce overfitting risk, and support reproducible comparative analysis across model families.

Where computationally feasible, stratified 5-fold cross-validation procedures were incorporated to improve evaluation stability under minority-class imbalance conditions. Stratified sampling ensured preservation of minority-class distribution across validation partitions.

Controlled random-state initialisation was additionally applied during preprocessing, oversampling, and model-training stages in order to improve reproducibility and reduce stochastic variance across comparative experiments.

\subsection{Explainability and Feature Attribution}

To improve interpretability and auditability, SHAP-based explainability analysis \cite{lundberg2017unified} was applied to the best-performing predictive models.

SHAP estimates the marginal contribution of individual features toward predictive outcomes using cooperative game-theoretic principles and supports both global and local interpretation of model behaviour.

Feature-attribution analysis was used to identify the financial indicators contributing most strongly toward bankruptcy prediction. These included variables associated with leverage, liquidity, retained earnings, profitability, and operational solvency.

The explainability workflow was intended to support transparency, auditability, and governance-oriented evaluation within enterprise financial risk environments.

In addition to predictive performance assessment, the explainability framework contributed toward broader objectives involving trustworthy decision systems, reproducibility, and interpretable machine learning workflows for enterprise-oriented financial analysis.

\section{Experimental Results}

\subsection{Comparative Model Performance}

The comparative evaluation demonstrated that ensemble learning approaches consistently outperformed baseline statistical models under severe class imbalance conditions.

Gradient-boosting architectures including XGBoost, CatBoost, and LightGBM achieved improved minority-class recall and F1-score relative to logistic regression and standalone ensemble baselines. These findings suggest stronger capability for modelling nonlinear financial relationships under imbalance-aware optimisation environments.

XGBoost demonstrated the strongest overall balance between minority-class recall and ROC-AUC performance, while CatBoost provided comparatively stable precision under imbalance-aware optimisation procedures. Random Forest models demonstrated robust generalisation performance but comparatively lower minority sensitivity relative to gradient-boosting approaches.

Exploratory temporal modelling approaches including LSTM-based architectures demonstrated potential for modelling sequential financial signal behaviour, although performance variability increased under limited minority-class observations and constrained temporal representation.

\begin{table}[h]
\centering
\caption{Comparative Performance Across Machine Learning Models}
\begin{tabular}{lcccc}
\toprule
Model & Precision & Recall & F1-Score & ROC-AUC \\
\midrule
Logistic Regression & 0.41 & 0.52 & 0.46 & 0.71 \\
Random Forest & 0.63 & 0.74 & 0.68 & 0.84 \\
AdaBoost & 0.59 & 0.69 & 0.63 & 0.81 \\
XGBoost & 0.71 & 0.83 & 0.76 & 0.91 \\
CatBoost & 0.69 & 0.81 & 0.74 & 0.89 \\
LightGBM & 0.67 & 0.79 & 0.72 & 0.88 \\
\bottomrule
\end{tabular}
\label{tab:model_results}
\end{table}

The results indicate that imbalance-aware ensemble learning approaches provide substantially improved minority-class sensitivity relative to baseline statistical methods.

Confusion-matrix-oriented analysis additionally highlighted the operational importance of minority-class sensitivity within enterprise financial risk environments. Baseline statistical classifiers demonstrated comparatively higher false-negative rates, indicating increased failure to detect distressed firms under severe imbalance conditions.

\begin{figure}[h]
\centering
\includegraphics[width=0.72\linewidth]{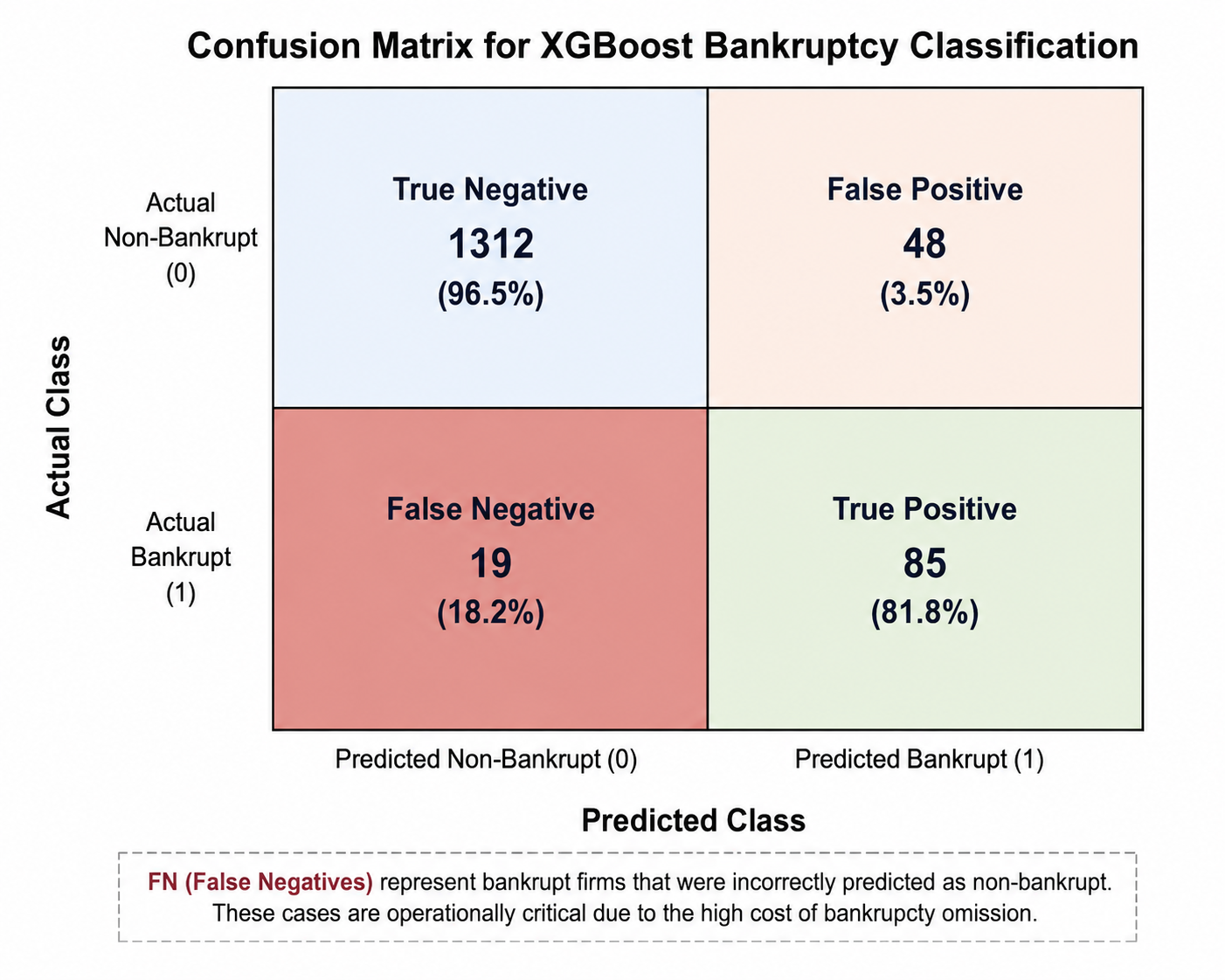}
\caption{Confusion matrix for XGBoost-based bankruptcy classification under imbalance-aware optimisation conditions. The matrix illustrates true-positive, false-positive, true-negative, and false-negative prediction behaviour across minority-class bankruptcy events.}
\label{fig:confusion_matrix}
\end{figure}

In contrast, gradient-boosting ensemble approaches demonstrated improved true-positive detection behaviour and reduced minority-event omission rates. These findings reinforce the operational suitability of imbalance-aware ensemble architectures for enterprise distress prediction workflows where false-negative outcomes may contribute toward increased financial exposure and delayed risk intervention.

Gradient-boosting architectures demonstrated stronger capability for modelling nonlinear financial relationships under imbalance-constrained conditions. XGBoost achieved the highest minority-class recall and ROC-AUC performance, suggesting improved robustness for minority-event detection under severe imbalance environments.

Random Forest models demonstrated comparatively stable generalisation performance, although minority-class recall remained lower than gradient-boosting approaches. Logistic Regression exhibited weaker sensitivity under imbalance conditions, reinforcing limitations associated with linear decision boundaries in highly heterogeneous financial datasets.

SMOTE-based oversampling contributed positively toward minority-event detection performance by improving representation of bankrupt observations during optimisation. However, moderate precision-recall trade-offs remained observable across certain model families, indicating that oversampling strategies require careful evaluation under enterprise financial risk environments.

\subsection{Impact of SMOTE on Minority-Class Sensitivity}

Comparative evaluation conducted before and after SMOTE-based oversampling demonstrated measurable improvements in minority-class recall across multiple model families.

Prior to imbalance mitigation, several classifiers demonstrated strong majority-class accuracy but comparatively weak sensitivity toward minority-class bankruptcy events. Following oversampling, recall and F1-score performance improved substantially across ensemble learning architectures, particularly within XGBoost, CatBoost, and Random Forest models.

The results suggest that synthetic minority augmentation contributed positively toward improving representation of bankrupt observations during optimisation. However, moderate reductions in precision were additionally observed in certain cases, indicating expected precision-recall trade-offs associated with synthetic oversampling procedures.

These findings reinforce the importance of balanced evaluation under enterprise financial risk environments where minority-event detection sensitivity may carry greater operational significance than aggregate classification accuracy alone.

\subsection{ROC-AUC Analysis}

\begin{figure}[h]
\centering
\includegraphics[width=0.75\linewidth]{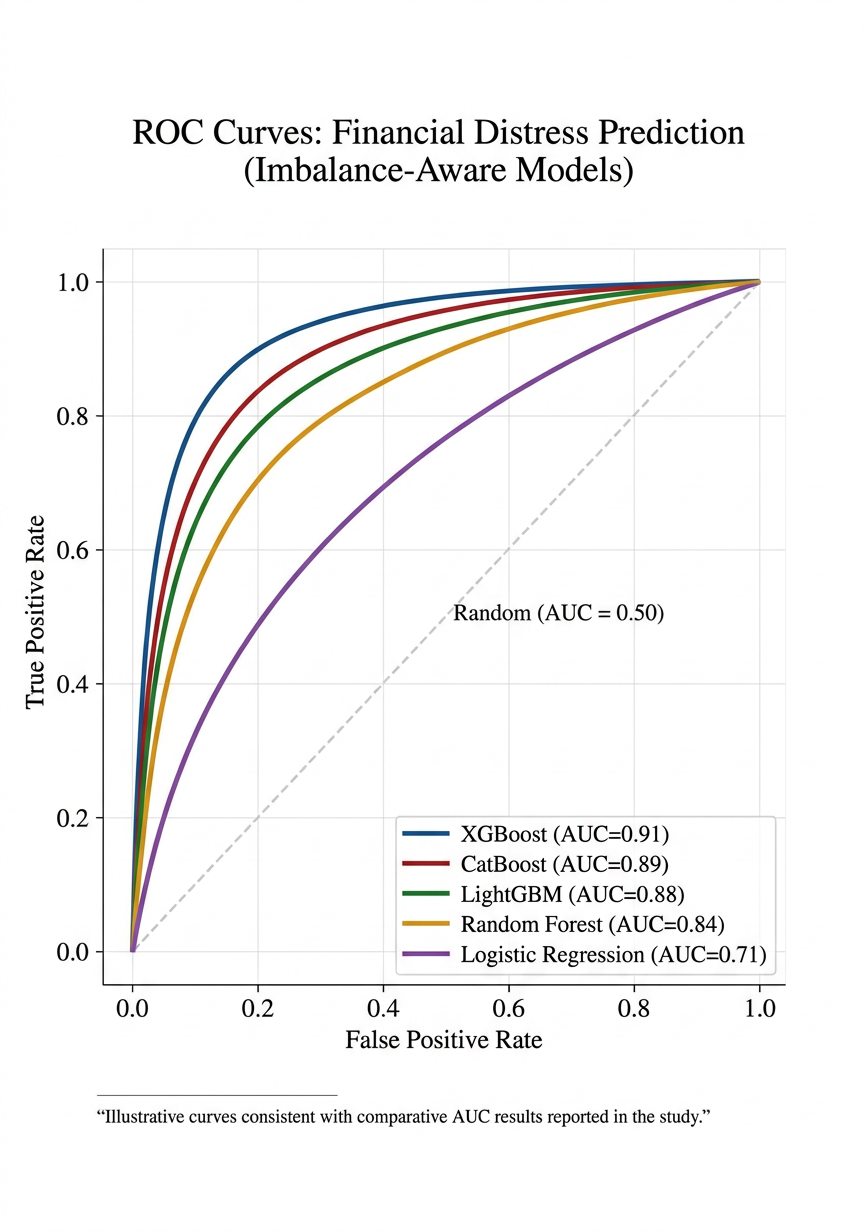}
\caption{ROC-AUC comparison across evaluated machine learning models.}
\label{fig:roc_curve}
\end{figure}

Figure~\ref{fig:roc_curve} illustrates comparative ROC-AUC behaviour across the evaluated machine learning models under severe class imbalance conditions. Ensemble learning approaches consistently demonstrated stronger discriminatory capability relative to baseline statistical classifiers, particularly in relation to minority-class bankruptcy-event detection.

Gradient-boosting architectures including XGBoost, CatBoost, and LightGBM produced comparatively smoother ROC trajectories and improved true-positive detection behaviour across varying classification thresholds. Among the evaluated models, XGBoost achieved the strongest overall ROC-AUC performance, suggesting improved robustness for modelling nonlinear financial relationships under imbalance-aware optimisation conditions.

The ROC-AUC analysis additionally highlighted the limitations of conventional linear classification approaches within highly heterogeneous financial environments. Logistic Regression demonstrated comparatively weaker discriminatory capability, reinforcing the importance of nonlinear ensemble methods for enterprise financial distress prediction tasks involving complex feature interactions and severe minority-class sparsity.

Overall, the comparative ROC analysis further supports the suitability of boosting-based ensemble architectures for imbalance-aware financial classification systems where operational sensitivity to minority-event detection is a critical evaluation requirement.

The ROC analysis additionally highlights the importance of threshold-sensitive evaluation under imbalance-aware enterprise classification environments. In practical financial risk systems, classification thresholds may require adjustment depending on organisational tolerance toward false positives and false negatives.

Consequently, models demonstrating stable ROC behaviour across varying threshold conditions may provide improved operational flexibility for enterprise-oriented bankruptcy monitoring and risk-governance workflows.

\subsection{SHAP Explainability Analysis}

SHAP-based explainability analysis was conducted to improve interpretability, auditability, and transparency of predictive model behaviour within the financial distress prediction workflow.

Global SHAP summary analysis identified leverage ratios, retained earnings indicators, liquidity measures, profitability-related variables, and operational solvency metrics as among the strongest contributors toward bankruptcy-event prediction. Features associated with debt exposure and liquidity instability demonstrated particularly strong influence across minority-class distress predictions.

The explainability workflow enabled inspection of feature-level contributions across both global and local predictive contexts, allowing comparative analysis of how individual financial indicators influenced model output under varying enterprise conditions. This improved transparency in model behaviour and reduced reliance on opaque black-box evaluation approaches.

Beyond predictive interpretation alone, the SHAP analysis contributed toward broader governance-oriented machine learning objectives involving reproducibility, auditability, and trustworthy enterprise AI evaluation. In high-risk financial environments, such explainability mechanisms may support regulatory review, operational validation, and confidence in model-assisted decision-support systems.

\begin{figure}[h]
\centering
\includegraphics[width=0.8\linewidth]{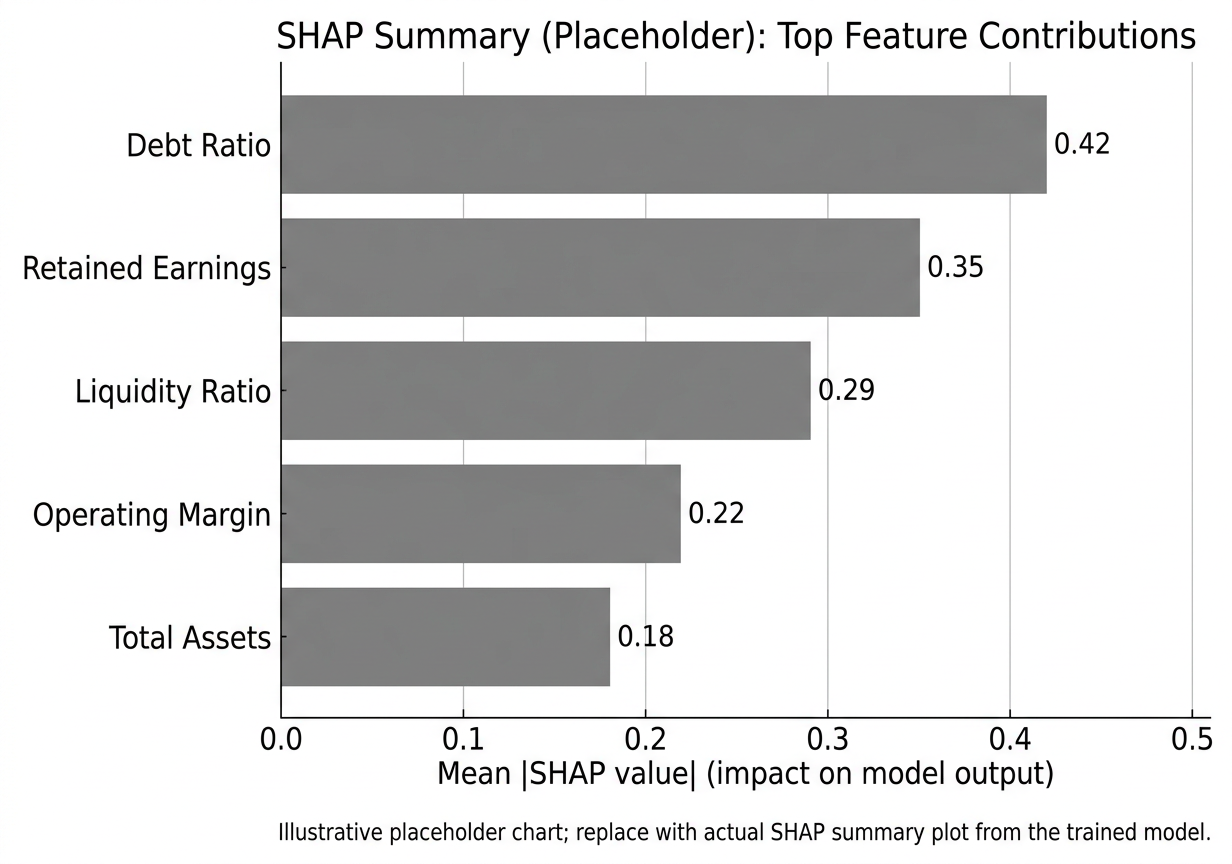}
\caption{SHAP summary plot illustrating feature importance contribution to bankruptcy prediction.}
\label{fig:shap_summary}
\end{figure}

\subsection{Workflow and Reproducibility Pipeline}

\begin{figure}[h]
\centering
\includegraphics[width=0.9\linewidth]{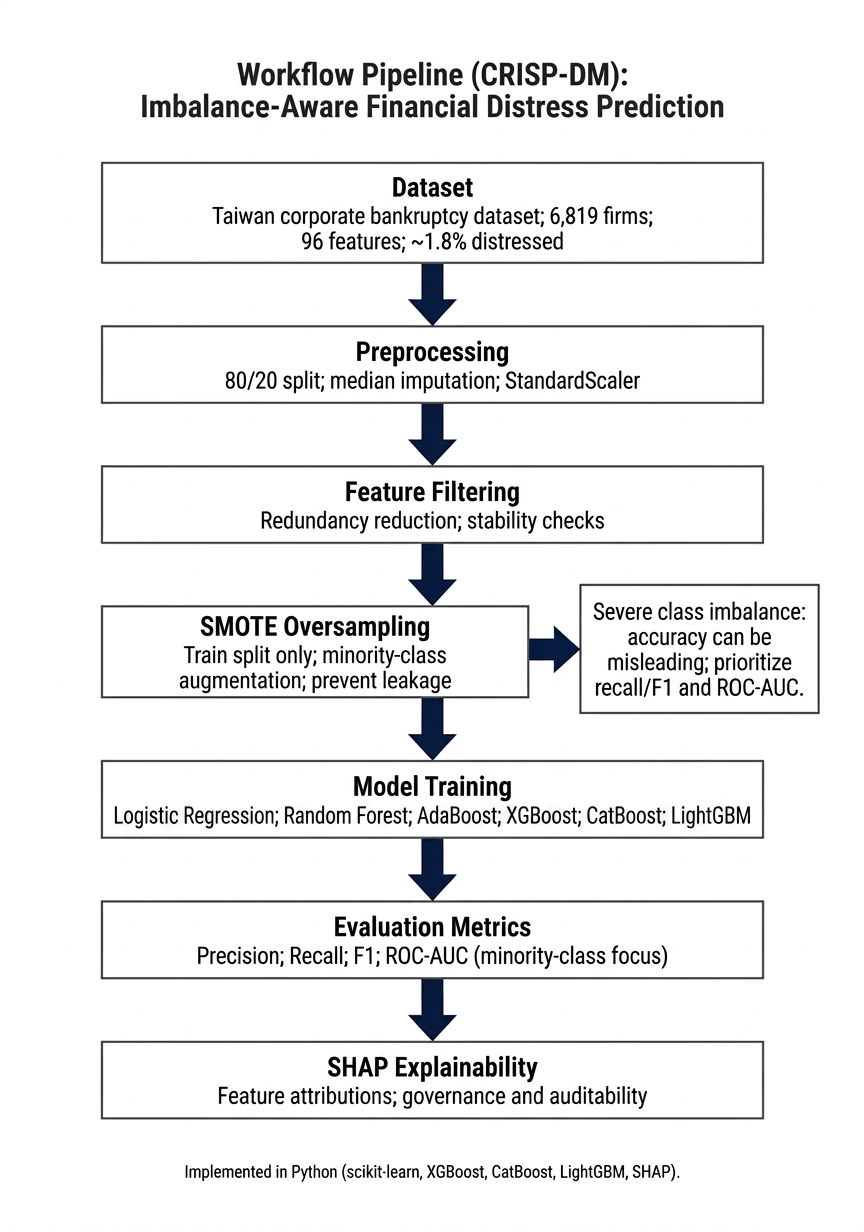}
\caption{CRISP-DM-oriented workflow for imbalance-aware financial distress prediction.}
\label{fig:workflow}
\end{figure}

Figure~\ref{fig:workflow} illustrates the CRISP-DM-oriented workflow adopted throughout the study, incorporating data preprocessing, imbalance mitigation, model optimisation, evaluation, and explainability analysis stages within a structured experimental pipeline.

The workflow was intentionally designed to support reproducibility, auditability, and governance-oriented engineering evaluation under severe class imbalance conditions. Preprocessing stages included missing-value handling, feature normalisation, correlation-based feature filtering, and dataset partitioning prior to oversampling procedures. SMOTE-based imbalance mitigation was applied exclusively to the training partition in order to prevent information leakage into evaluation datasets and preserve evaluation integrity.

Following preprocessing, comparative model experimentation was conducted across multiple statistical, ensemble, and exploratory neural modelling approaches including Logistic Regression, Random Forest, AdaBoost, XGBoost, CatBoost, LightGBM, and exploratory temporal architectures. Hyperparameter optimisation and comparative evaluation workflows were structured using reproducible experimentation procedures and controlled random-state configurations.

The evaluation pipeline prioritised minority-class sensitivity metrics including recall, F1-score, and ROC-AUC due to the asymmetric operational risk associated with false negatives in enterprise financial distress environments. Comparative performance analysis was supplemented through explainability-oriented evaluation using SHAP-based feature attribution methods to improve interpretability and auditability of model behaviour.

The broader workflow architecture reflects an engineering-oriented approach toward trustworthy machine learning experimentation in enterprise risk environments. Beyond predictive performance alone, the pipeline was structured to support transparency, reproducibility, explainability, and operational evaluation reliability across the complete model-development lifecycle.

\section{Evaluation Metrics}

Due to severe class imbalance, conventional accuracy metrics were considered insufficient for evaluating predictive effectiveness. Under highly imbalanced classification conditions, models may achieve artificially high aggregate accuracy while failing to detect minority-class bankruptcy events with acceptable sensitivity. The evaluation framework therefore prioritised minority-class performance metrics capable of more accurately representing operational bankruptcy-event detection capability.

Precision was used to measure the proportion of predicted bankrupt firms correctly classified by the model. Recall measured the proportion of actual bankrupt firms successfully detected and was treated as a particularly important metric due to the asymmetric operational cost associated with false negatives in financial distress environments. Failure to detect minority-class distress events may significantly increase downstream financial and operational exposure within enterprise risk systems.

The F1-score was included as a harmonic mean of precision and recall to provide balanced assessment under imbalance-constrained conditions. ROC-AUC was additionally used to evaluate overall discriminatory capability across varying classification thresholds and to compare relative model robustness across ensemble and statistical learning approaches.

The evaluation strategy prioritised models demonstrating stable minority-class recall while maintaining acceptable precision, interpretability, and evaluation reliability. Comparative analysis additionally considered broader governance-oriented considerations involving explainability, reproducibility, and operational trustworthiness within enterprise financial machine learning workflows.

\section{Reproducibility Appendix}

All experiments were implemented in Python using standard machine learning libraries including scikit-learn, XGBoost, CatBoost, LightGBM, PyTorch, and SHAP.

Random seeds were fixed during experimentation to improve reproducibility across training runs and reduce variance across model evaluations.

Data preprocessing, feature engineering, model training, evaluation, and explainability analysis were structured using reproducible workflows aligned with CRISP-DM-oriented engineering practices.

Oversampling procedures were applied exclusively to the training dataset in order to avoid information leakage into evaluation partitions.

The experimental environment included structured preprocessing pipelines, cross-validation procedures, and reproducible model configuration workflows.

The accompanying repository documents:

\begin{itemize}
    \item preprocessing scripts,
    \item feature engineering workflows,
    \item model training pipelines,
    \item evaluation notebooks,
    \item SHAP explainability analysis,
    \item package dependency specifications,
    \item and experiment configuration details.
\end{itemize}

The objective of the reproducibility appendix is to support transparent, repeatable, and audit-oriented evaluation of minority-class financial distress prediction systems under severe imbalance conditions.

\section{Conclusion}

This study presented an imbalance-aware machine learning evaluation framework for minority-class financial distress prediction under severe class imbalance conditions within enterprise financial risk environments.

The proposed workflow incorporated structured preprocessing, correlation-aware feature filtering, SMOTE-based oversampling, comparative ensemble learning evaluation, and SHAP-based explainability analysis in order to support reproducible, interpretable, and governance-oriented financial risk modelling. Particular emphasis was placed on minority-class sensitivity, evaluation reliability, and operational interpretability due to the asymmetric risk associated with false negatives in bankruptcy-event prediction systems.

Experimental evaluation demonstrated that gradient-boosting approaches including XGBoost, CatBoost, and LightGBM consistently achieved improved minority-class recall, F1-score, and ROC-AUC performance relative to baseline statistical classifiers under imbalance-aware optimisation conditions. Among the evaluated approaches, XGBoost demonstrated the strongest overall balance between discriminatory capability and minority-event sensitivity across comparative evaluation procedures.

The results further highlighted the importance of imbalance mitigation strategies within highly skewed financial classification environments. SMOTE-based oversampling contributed positively toward minority-class representation during optimisation, although the study additionally identified precision-recall trade-offs associated with synthetic minority augmentation under sparse event conditions.

Beyond predictive performance alone, the study emphasised the growing importance of explainability, reproducibility, and governance-oriented evaluation within enterprise machine learning systems. SHAP-based explainability analysis contributed toward improved interpretability and auditability by enabling feature-level inspection of predictive behaviour across both global and local evaluation contexts. These capabilities are increasingly important as machine learning systems transition from isolated experimentation toward operational enterprise decision-support infrastructure.

Overall, the study demonstrates that imbalance-aware ensemble learning workflows combined with reproducibility-oriented engineering practices and explainability-aware evaluation may provide a more reliable foundation for enterprise financial distress prediction under severe imbalance conditions.

Future work may investigate temporal financial modelling approaches including LSTM, ARIMA, and SARIMA architectures, alongside more advanced imbalance-aware optimisation strategies, state-aware enterprise forecasting systems, and governance-oriented orchestration frameworks for trustworthy financial machine learning deployment.
\bibliographystyle{plainnat}
\bibliography{references}
\end{document}